\documentclass[letterpaper, 10 pt, conference]{ieeeconf}  % Comment this line out if you need a4paper

\IEEEoverridecommandlockouts                              % This command is only needed if 
\usepackage{hyperref}
\hypersetup{bookmarksopen,bookmarksnumbered,
pdfpagemode=UseOutlines,
colorlinks=true,
linkcolor=blue,
anchorcolor=blue,
citecolor=blue,
filecolor=blue,
menucolor=blue,
urlcolor=blue
}
\usepackage{upgreek}
\usepackage{amsfonts,amssymb}
\usepackage{bm}
\usepackage{bbm}

\usepackage{amsthm} % for new theorem
\usepackage{thmtools}
\usepackage{mathtools}
\usepackage{xspace}
\usepackage{units}
\usepackage{booktabs} %for table rulers
\usepackage{tabulary}
\usepackage[usenames,dvipsnames,table]{xcolor} %Options expand the named colours
\usepackage{diagbox}
\usepackage[ruled,vlined,linesnumbered]{algorithm2e}  %updated version of algorithm2e.sty is in ./ folder.
\usepackage{parskip}
\SetKwComment{Comment}{$\triangleright$\ }{}
\usepackage{ifthen,version}
\usepackage{soul}
\usepackage{csquotes}
\usepackage{microtype}      % microtypography
\usepackage{lipsum}
\usepackage{tikz}
\let\labelindent\relax
\usepackage{todonotes}
\usepackage{enumitem}
\usepackage{cite}

\newcommand{\xxnote}[3]{}
\ifx\hidenotes\undefined
  \usepackage{color}
  \renewcommand{\xxnote}[3]{\color{#2}{#1: #3}}
\fi

\usepackage{multirow}
\usepackage{pbox}

\newtheoremstyle{hypstyle}
{3pt} % Space above
{3pt} % Space below
{\itshape} % Body font
{} % Indent amount
{\bfseries} % Theorem head font
{.} % Punctuation after theorem head
{.5em} % Space after theorem head
{} % Theorem head spec (can be left empty, meaning `normal')

\theoremstyle{hypstyle}

\DeclareMathOperator*{\argmin}{arg\,min}

\newcommand{\argminprob}[1]{\underset{#1}{\argmin}}
\newcommand{\norm}[2]{\left|\left| #1 \right|\right|_{#2}}

\newcommand{\real}[0]{\mathbb{R}}

\newcommand{\bbm}{\begin{bmatrix}}
\newcommand{\ebm}{\end{bmatrix}}

\newcommand{\Path}[1]{\xi_{#1}}
\newcommand{\cost}[0]{J}
\newcommand{\costFn}[1]{\cost \left( #1 \right)}

\newcommand{\costShot}[0]{\cost_\mathrm{shot}}
\newcommand{\costFnShot}[1]{\costShot \left( #1 \right)}

\newcommand{\costSmooth}[0]{\cost_\mathrm{smooth}}
\newcommand{\costFnSmooth}[1]{\costSmooth \left( #1 \right)}

\newcommand{\costOcc}[0]{\cost_\mathrm{occ}}
\newcommand{\costFnOcc}[1]{\costOcc \left( #1 \right)}

\newcommand{\costObs}[0]{\cost_\mathrm{obs}}
\newcommand{\costFnObs}[1]{\costObs \left( #1 \right)}

\newcommand{\map}[0]{\mathcal{M}}
\newcommand{\grid}[0]{\mathcal{G}}

\graphicspath{{figs/}}

\title{\LARGE \bf
Towards a Robust Aerial Cinematography Platform: \\Localizing and Tracking Moving Targets in Unstructured Environments
}
\author{Rogerio Bonatti $^{1}$, Cherie Ho $^{1}$, Wenshan Wang $^{1}$, Sanjiban Choudhury $^{2}$ and Sebastian Scherer $^{1}$% <-this % stops a space
\thanks{*Research presented in this paper was funded by Yamaha Motor Co., Ltd.}% <-this % stops a space
\thanks{$^{1}$R. Bonatti, C. Ho, W. Wang, S. Scherer belong to The Robotics Institute, School of Computer Science,
        Carnegie Mellon University, Pittsburgh PA
        {\tt\small \{rbonatti,cherieh,wenshanw,basti\}@cs.cmu.edu}}%
\thanks{$^{2}$S. Choudhury is with the School of Computer Science and Engineering of the University of Washington, Seattle, WA
        {\tt\small sanjibac@cs.uw.edu}}%
}

\begin{document}

\maketitle
\thispagestyle{empty}
\pagestyle{empty}

\begin{abstract}

% \rbnote{re-write to convey the three main points that we have in this paper, and to reflect our experiments}

% \rbnote{a few things to consider for abstract: - Context: What is the context of your work, what is the start of the art 
% - Need: What is the lack of the start of the art 
% - Task: What is the task you want to address in your work (should be 1 sentence) 
% - Object: How do you plan to address your task? 
% - Results: 
% - Conclusions: What are your expected conclusions?}

The use of drones for aerial cinematography has revolutionized several applications and industries that require live and dynamic camera viewpoints such as entertainment, sports, and security. However, safely controlling a drone while filming a moving target usually requires multiple expert human operators; hence the need for an autonomous cinematographer. 
Current approaches have severe real-life limitations such as requiring fully scripted scenes, high-precision motion-capture systems or GPS tags to localize targets, and prior maps of the environment to avoid obstacles and plan for occlusion.

In this work, we overcome such limitations and propose a complete system for aerial cinematography that combines: (1) a vision-based algorithm for target localization; (2) a real-time incremental 3D signed-distance map algorithm for occlusion and safety computation; and (3) a real-time camera motion planner that optimizes smoothness, collisions, occlusions and artistic guidelines. We evaluate robustness and real-time performance in series of field experiments and simulations by tracking dynamic targets moving through unknown, unstructured environments. Finally, we verify that despite removing previous limitations, our system achieves state-of-the-art performance.  

\end{abstract}

% !TEX root = ../root.tex

%%%%%%%%%%%%%%%%%%%%%%%%%%%%%%%%%%%%%%%%%%%%%%%%%%%%%%%%%%%%%%%%%%%%%%%%%%%%%%%%
\section{Introduction}

% \rbnote{re-draw main figure we can go in 2-column format to show: a) system cartoon that includes mapping and person detection besides planning and actor forecast b) third-person view of scene from spark with a sub-view of the DJI cam view. we can highlight important things like make bounding boxes, heading estimation ,etc like cover pic from icra paper, and c) RVIZ view of planner trajectory, environment poincloud (accumulated), and actor movement history including arrows for estimated odometry}

% \rbnote{we could also make this figure as 4 sub-figures within 1 single column of the paper. if so, we would have a) cartoon, b) spark 3-rd person view c) DJI camera view d) RVIZ view. try both approaches and see how it looks, but prob. 2-colum is more appealing}

\begin{figure*}[t]
    \center
    \includegraphics[width=1.0\textwidth]{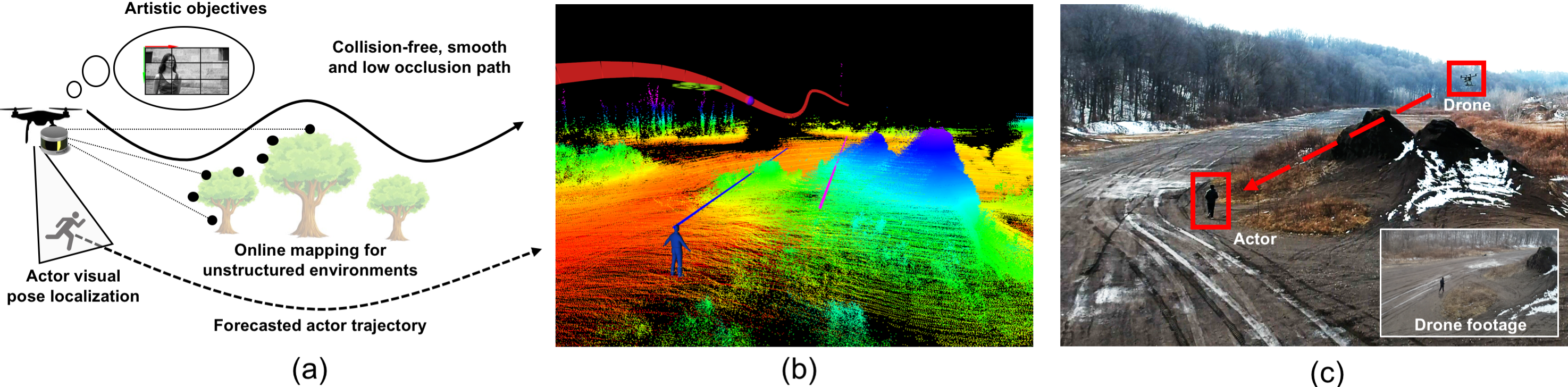}
    \caption{Aerial cinematographer: a) The UAV forecasts the actor's motion using camera-based localization, maps the environment with a LiDAR, reasons about artistic guidelines, and plans a smooth, collision-free trajectory while avoiding occlusions. b) Accumulated point cloud during field test overlaid with actor's motion forecast (blue), desired cinematography guideline (pink), and optimized trajectory (red). c) Third-person view of scene and final drone image. 
    \label{fig:intro}}
\end{figure*}

% \scnote{Begin with a sense of urgency. Make sure the sentences are simple and have a clear logical progression. Each paragraph should not be more than a few sentences. Strive to be done with intro in ~1 page.}

% \scnote{
% P1: What is the problem we want to solve? We want an UAV to film actors that are moving unpredictably through an outdoors, unstructured environment. 
% This is challenging for even the most experienced of pilots due to the attention and effort needed.}\rbnote{usually requires multiple professional pilots; unscripted scenes (makes a lot of difference if actor follows pre-defined route)}

In this paper, we address the problem of autonomous cinematography using unmanned aerial vehicles (UAVs). 
Specifically, we focus on scenarios where an UAV must film an actor moving through an unknown environment at high speeds, in an unscripted manner.
Filming dynamic actors among clutter is extremely challenging, even for experienced pilots. It takes high attention and effort to simultaneously predict how the scene is going to evolve, control the UAV, avoid obstacles and reach desired viewpoints. 
% In practice, critical applications require two or more pilots per UAV to guarantee good performance. \scnote{redundant?}
Towards solving this problem, we present a complete system that can autonomously handle the real-life constraints involved in aerial cinematography: tracking the actor, mapping out the surrounding terrain and planning maneuvers to capture high quality, artistic shots. 
% \scnote{why not skydio? }

% \scnote{
% P2: Tease out the nuances of the problem. 
% What should the UAV take into consideration? The UAV must make local predictions of how the actor will move, compute desired viewpoints that are free of occlusion and plan to visit them in a safe manner. 
% What is the central challenge? The environment is unknown and must be mapped on the fly. The predictions are uncertain. Planning must be done online. 
% } 

% \rbnote{plus we must localize the actor pose and forecast future motion -- also big challange. this way we cover all 3 big sub-systems in intro (plan, vision, map)}

% \scnote{This paragraph seems fishy - you are almost saying the UAV must do whatever your approach is. This will piss off a reviewer. Instead, say things like - the UAV must remain safe as it flies through new environments}

Consider the typical filming scenario in Fig~\ref{fig:intro}. The UAV must accomplish a number of tasks. First, it must estimate the actor's pose using an onboard camera and forecast their future motion. The pose estimation should be robust to changing viewpoints, backgrounds and lighting conditions. Accurate forecasting is key for anticipating events which require changing camera viewpoints.  
% In a typical aerial filming scenario, as shown in Fig~\ref{fig:intro}, the system must accomplish a number of tasks in real-time, under a limited onboard computational budget. First, it must be able to detect the target and predict their motion.  
% handle several tasks in real-time, under a limited onboard computational budget. First, since the UAV is tracking a target, it must use sensors, such as cameras, to visually localize the actor and predict their future motion in order to enable deliberative, and not just reactive, maneuvers. 
% Second, the vehicle must map the surrounding terrain, identifying occupied, free and unknown space \scnote{No,it must not. It must remain safe. That in turn demands reasoning about uncertainty - dont need to go into details of approach}. \chnote{Second, the vehicle must remain safe in an unmapped environment,...}
Secondly, the UAV must remain safe as it flies through through new environments. Safety requires explicit modelling of environmental uncertainty.
%Second, the vehicle must remain safe as it flies through through new environments, reasoning about its own uncertainties and about unknown areas of the map.
Finally, the UAV must capture high quality videos which require maximizing a set of artistic guidelines. 
The key challenge is that all these tasks must be done in real-time under limited onboard computational resources.

There is a rich history of work in autonomous aerial filming that tackles parts of the challenges. 
For instance, several works focus on artistic guidelines~\cite{joubert2016towards,nageli2017real,galvane2017automated,galvane2018directing} but often rely on perfect actor localization through high-precision RTK GNSS or motion-capture systems.
Additionally, while the majority of work in the area deals with collisions between UAV and actors\cite{nageli2017real,joubert2016towards,huang2018act}, the environment is not factored in.
 %but ignores the environment altogether \cite{huang2018act,nageli2017real,joubert2016towards}.
% We also see works focused on offline trajectory generation, mainly \cite{gebhardt2016airways,xie2018creating,joubert2016towards,roberts2016generating}
While there are several successful commercial products, they too have certain limitations to either low speed and low clutter regimes  (e.g. DJI Mavic~\cite{mavic}) or shorter planning horizons (e.g. Skydio R1~\cite{skydio2018}). 
%Commercial products also present limitations such as limited obstacle avoidance capabilities at low speeds and simple scenarios (\textit{e.g.} DJI Mavic \cite{mavic}) and relatively short planning time horizons (\textit{e.g.} Skydio R1 \cite{skydio2018}), which can limit the quality of obstacle and occlusions avoidance maneuvers.
Even our previous work~\cite{bonatti2018autonomous}, despite handling environmental occlusions and collisions, assumes a prior elevation map and uses GPS to localize the actor. 
%Prior maps are invalidated over time as the environment changes. 
Such simplifications impose restrictions on the diversity of real-life scenarios that these systems can handle.

We address these challenges by building upon previous work that formulates the problem as an efficient real-time trajectory optimization~\cite{bonatti2018autonomous}. In this work we make a key observation: we don't need prior ground-truth information about the scene; our onboard sensors suffice to attain good performance. However, sensor data is noisy and needs to be processed in real-time; therefore we develop robust and efficient algorithms. To localize the actor, we use a visual tracking system. To map the environment, we use a long-range LiDAR and process it incrementally to build a signed distance field of the environment. Combining both methods, we can plan over long horizons in unknown environments to film fast dynamic actors according to artistic guidelines. In summary, our main contributions in this paper are threefold:

\begin{enumerate}
  \item We develop an incremental signed distance transform algorithm for large-scale real-time environment mapping (Section~\ref{subsec:approach_mapping});
  \item We develop a complete system for autonomous cinematography that includes visual actor localization, online mapping, and efficient trajectory optimization that can deal with noisy measurements (Section~\ref{sec:approach});
  \item We offer extensive quantitative and qualitative performance evaluations of our system both in simulation and field tests, while also comparing performance changes with scenarios with full map and actor knowledge (Section~\ref{sec:experiments}).
\end{enumerate}

\section{Problem Formulation}
\label{sec:problem_formulation}
% \vspace{-2mm}

% \rbnote{make more complete}

% \begin{figure}[t]
%     \center
%     \includegraphics[scale=.36]{figs/cover_pic_blocks}
%     \caption{Side shot of a person filmed by the drone cinematographer: a) Time-lapse of actor and drone. b) Trajectory generation, where the actor's motion forecast is in purple, the desired cinematography position is in blue, and the optimized trajectory is shown in red. c) Resulting drone camera image}
%     \label{fig:cover_pic_blocks}
% \end{figure}		

% \rbnote{define problem: what exactly we're trying to solve in this paper mathematically. we're trying to film the actor following artistic principles (same things as written in ISER), trying to minimize a cost function with multiple costs. now the addition for IROS is that the cost functions for obstacle avoidance and occlusion depend on our distance transform of environment $\mathbb{E}$ that we capture online, and depends on the localization of the actor $\Path{a}$ that we estimate online as well (instead of being ground truth like previous work) }

The overall task is to control a UAV to film an actor who is moving through an unknown environment. We formulate this as a trajectory optimization problem where the cost function measures shot quality, environmental occlusion of the actor, jerkiness of motion and safety. This cost function depends on the environment and the actor, both of which must be sensed on-the-fly. The changing nature of environment and actor trajectory also demands re-planning at a high frequency. 

% \chnote{Suggested: Let $\Path{} : [0,t_f] \rightarrow \real^3  \times SO(2)$ be the general trajectory form over time horizon $t_f$, i.e. $\Path{}(t) = \{x(t), y(t), z(t), \psi(t)\}$, with $\psi$ is heading (yaw) angle. Let $\Path{q}$ be the trajectory of the UAV we are computing, and $\Path{a}$ be the trajectory of the actor. (The state of the actor...)}
Let $\Path{q} : [0,t_f] \rightarrow \real^3  \times SO(2)$ be the trajectory of the UAV, i.e., $\Path{q}(t) = \{x(t), y(t), z(t), \psi_{q}(t)\}$ . 
% \chnote{define $\psi(t)$} 
Let $\Path{a} : [0,t_f] \rightarrow \real^3  \times SO(2)$ be the trajectory of the actor, $\Path{a}(t) = \{x(t), y(t), z(t), \psi_{a}(t)\}$.
% \chnote{does $\Path{a}(t)$ also contain $x,y,z,\psi$}
The state of the actor, as sensed by onboard cameras, is fed into a prediction module that computes $\Path{a}$ (Section~\ref{subsec:approach_vision}). 

Let grid $\grid: \real^3 \rightarrow \real$ be a voxel occupancy grid that maps every point in space to a probability of occupancy. Let $\map : \real^3 \rightarrow \real$ be the signed distance values of a point to the nearest obstacle. Positive sign is for points in free space, and negative sign is for points either in occupied or unknown space, which we assume to be potentially inside an obstacle. The UAV senses the environment with the onboard LiDAR, updates grid $\grid$, and then updates $\map$ (Section~\ref{subsec:approach_mapping}).

We briefly touch upon the four components of the cost function $\costFn{\Path{q}}$ (refer to Section~\ref{subsec:approach_planning} for mathematical expressions). The objective is to minimize $\costFn{\Path{q}}$ subject to initial boundary constraints $\Path{q}(0)$. 

\begin{enumerate}
	\item \emph{Smoothness} $\costFnSmooth{\Path{q}}$: Penalizes jerky motions that may lead to camera blur and unstable flight;
	\item \emph{Shot quality} $\costFnShot{\Path{q},\Path{a}}$: Penalizes poor viewpoint angles and scales that deviate from the artistic guidelines
	\item \emph{Safety} $\costFnObs{\Path{q},\map}$: Penalizes proximity to obstacles that are unsafe for the UAV.
	\item \emph{Occlusion} $\costFnOcc{\Path{q},\Path{a},\map}$: Penalizes occlusion of the actor by obstacles in the environment.
\end{enumerate}

\begin{equation}
\begin{aligned}
\label{eq:main_cost}
\costFn{\Path{q}} &=  \begin{bmatrix}
       1 & \lambda_1 & \lambda_2 & \lambda_3
     \end{bmatrix} 
\begin{bmatrix}
       \costFnSmooth{\Path{q}} \\
       \costFnShot{\Path{q},\Path{a}} \\
       \costFnObs{\Path{q},\map} \\
       \costFnOcc{\Path{q},\Path{a},\map} 
     \end{bmatrix} \\
%\costFnSmooth{\Path{q}} + \lambda_1 \costFnObs{\Path{q}} + \lambda_2 \costFnOcc{\Path{q},\Path{a}} + \lambda_3 \costFnShot{\Path{q},\Path{a}} \\
\Path{q}^* &= \argminprob{} \quad \costFn{\Path{q}}, \quad \text{s.t.} \; \Path{q}(0) = \{x_0,y_0,z_0,\psi_0\}
\end{aligned}
\end{equation}
% \chnote{choice of $\lambda$?}
The solution $\Path{q}^*$ is then tracked by the UAV.
\section{Related Work} 
\label{sec:related_work}

\paragraph{Virtual cinematography} Camera control in virtual cinematography has been extensively examined by the computer graphics community, as reviewed by \cite{christie2008camera}. These methods tend to reason about the utility of a viewpoint in isolation, following artistic principles and composition rules \cite{arijon1976grammar,bowen2013grammar} and employ either optimization-based approaches to find good viewpoints, or reactive approaches to track the virtual actor. The focus is typically on through-the-lens control where a virtual camera is manipulated while maintaining focus on certain image features~\cite{gleicher1992through,drucker1994intelligent,lino2011director,lino2015intuitive}. 
However, virtual cinematography is free of several real-world limitations such as robot physics constraints and assumes full map knowledge.

\paragraph{Autonomous aerial cinematography}
Several contributions on aerial cinematography focus on keyframe navigation. \cite{roberts2016generating,joubert2015interactive,gebhardt2018optimizing,gebhardt2016airways,xie2018creating} provide user interface tools for re-timing and connecting static aerial viewpoints for dynamically feasible and visually pleasing trajectories. \cite{lan2017xpose} use key-frames defined on the image itself instead of world coordinates.

Other works focus on tracking dynamic targets, and employ a diverse set of techniques for actor localization and navigation. For example, \cite{huang2018act,huang2018through} detect the skeleton of targets from visual input, while others approaches rely on off-board actor localization methods from either motion-capture systems or GPS sensors \cite{joubert2016towards,galvane2017automated,nageli2017real,galvane2018directing,bonatti2018autonomous}. These approaches have a varying level of complexity: \cite{bonatti2018autonomous,galvane2018directing} can avoid obstacles and occlusions with the environment and with actors, while other approaches only handle collisions and occlusions caused by actors. We also observe distinct trajectory generation methods randing from trajectory optimization to search-based planners. In Table~\ref{tab:related_work} we summarize different contributions, also differentiating onboard versus off-board computing systems. It is important to notice that prior to our current work, none of the previous approaches provided a solution for online environment mapping.

\begin{table}[t]
\begin{center}
\caption{Comparison of dynamic aerial cinematography systems}
\begin{tabular}{l|llllll}
\label{tab:related_work}

 \textbf{\pbox{0cm}{Ref}}    & \textbf{\pbox{0.6cm}{Online\\map\\}}  & \textbf{\pbox{0.6cm}{Actor\\localiz.\\}} & \textbf{\pbox{0.9cm}{Onboard\\comp.\\}}    & \textbf{\pbox{0cm}{Avoids \\occl.\\}}   & \textbf{\pbox{0cm}{Avoids\\obst.\\}} & \textbf{\pbox{0.6cm}{Online\\plan\\}} \\
 \hline
 \cite{galvane2017automated} &  \cellcolor{red!25} $\times$          & \cellcolor{red!25} $\times$              & \cellcolor{red!25} $\times$               & \cellcolor{red!25} $\times$              & \cellcolor{red!25} $\times$          & \cellcolor{green!25} \checkmark       \\
 \cite{joubert2016towards}   &  \cellcolor{red!25} $\times$          & \cellcolor{red!25} $\times$              & \cellcolor{red!25} $\times$               & \cellcolor{red!25} $\times$              & \cellcolor{yellow!25}  Actor     & \cellcolor{green!25} \checkmark       \\  
 \cite{nageli2017real}       &  \cellcolor{red!25} $\times$          & \cellcolor{red!25} $\times$              & \cellcolor{red!25} $\times$               &\cellcolor{yellow!25} Actor         & \cellcolor{yellow!25} Actor     & \cellcolor{green!25} \checkmark       \\
 \cite{galvane2018directing} &  \cellcolor{red!25} $\times$          & \cellcolor{red!25} $\times$              & \cellcolor{red!25} $\times$               &\cellcolor{green!25} \checkmark           & \cellcolor{green!25} \checkmark       & \cellcolor{green!25} \checkmark       \\
 
 \cite{huang2018through}     &  \cellcolor{red!25} $\times$          & \cellcolor{green!25} \checkmark          & \cellcolor{green!25} \checkmark           & \cellcolor{red!25} $\times$              & \cellcolor{yellow!25} Actor     & \cellcolor{green!25} \checkmark       \\
 \cite{huang2018act}         &  \cellcolor{red!25} $\times$          & \cellcolor{green!25} $\checkmark$        & \cellcolor{green!25} \checkmark           &\cellcolor{red!25} $\times$               & \cellcolor{yellow!25} Actor     & \cellcolor{green!25} \checkmark       \\
 \cite{bonatti2018autonomous}&  \cellcolor{red!25} $\times$          & \cellcolor{yellow!25} Vision              & \cellcolor{green!25} \checkmark           &\cellcolor{green!25} \checkmark           & \cellcolor{green!25} \checkmark       & \cellcolor{green!25} \checkmark       \\
 \textbf{Ours}               & \cellcolor{green!25} \checkmark       & \cellcolor{green!25} \checkmark          & \cellcolor{green!25} \checkmark           &  \cellcolor{green!25} \checkmark         & \cellcolor{green!25} \checkmark       & \cellcolor{green!25} \checkmark       \\

\end{tabular}
\end{center}
\end{table}

\paragraph{Online environment mapping}
Dealing with imperfect representations of the world becomes a bottleneck for viewpoint optimization in physical environments. As the world is sensed online, it is usually incrementally mapped using voxel occupancy maps~\cite{thrun2005probabilistic}. To evaluate a viewpoint, methods typically raycast on such maps, which can be very expensive~\cite{isler2016information,Charrow-RSS-15}. Recent advances in mapping have led to better representations that can incrementally compute the \emph{truncated signed distance field (TSDF)}~\cite{newcombe2011kinectfusion,klingensmith2015chisel}, i.e. return the distance and gradient to nearest object surface for a query. TSDFs are a suitable abstraction layer for planning approaches and have already been used to efficiently compute collision-free trajectories for  UAVs~\cite{oleynikova2016voxblox,cover2013sparse}. 
% \chnote{For mapping, we use TSDFs, which has already been used to efficiently compute collision-free trajectories for  UAVs~\cite{oleynikova2016voxblox,cover2013sparse}.}

% \rbnote{besides the table that compares work specifically on aerial cinematography, list related work on 1. motion planning for UAVs in general, 2. artistic stuff (the 2 books), 3. online mapping creation (voxblox, signed distance, old stuff from basti on distance trnasform), 4. bbox detection and heading estimation (copy from ICRA paper)}

\paragraph{Visual target state estimation}
% \rbnote{CH: finish this part, be very concise}
Accurate object state estimation with monocular cameras is critical to many robot applications. Deep networks have shown success in detecting objects \cite{yolov3,faster_rcnn} and estimating 3D heading \cite{raza2018appearance,li2014} with several efficient architectures developed specifically for mobile applications \cite{howard2017mobilenets, Zhang_2018_CVPR}. However, many models do not generalize well to other tasks (e.g., aerial filming) due to data mismatch in terms of angles and scales. Our recent work in semi-supervised learning shows promise in increasing model generalizability with little labeled data by leveraging temporal continuity in training videos \cite{wang2019heading}. 
Our work exploits synergies at the confluence of several domains of research to develop an aerial cinematography platform that can follow dynamic targets in unknown and unstructured environments, as detailed next in our approach.
% !TEX root = ../root.tex

\section{Approach} 
\label{sec:approach}

\begin{figure}[t]
    \centering
    \includegraphics[width=0.45\textwidth]{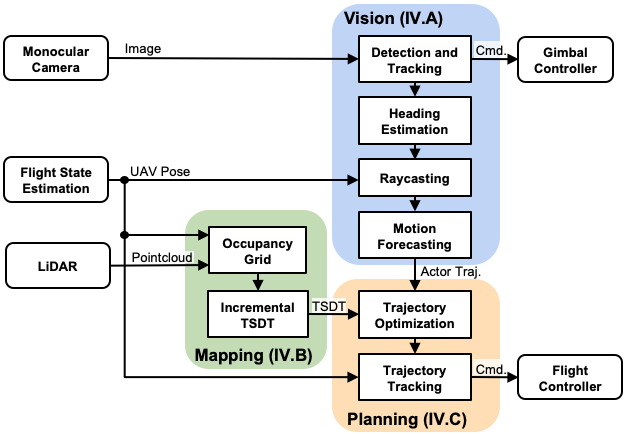}
    \caption{System architecture. Vision subsystem controls camera's orientation and forecasts the actor trajectory $\Path{a}$ using  monocular images. Mapping receives LiDAR pointclouds to incrementally calculate a truncated signed distance transform (TSDT). Planning uses the map, the drone's state estimation and actor's forecast to generate trajectories for the flight controller.}
    \label{fig:system2}
\end{figure}

We now detail our approach for each sub-system of the aerial cinematography platform. At a high-level, three main sub-systems operate together: (A) Vision, required for localizing the target's position and orientation and for recording the final UAV's footage; (B) Mapping, required for creating an environment representation; and (C) Planning, which combines the actor's pose and the environment to calculate the UAV trajectory. Fig.~\ref{fig:system2} shows a system diagram.

% !TEX root = ../root.tex

\subsection{Vision sub-system}
\label{subsec:approach_vision}

\begin{figure}[b]
    \center
    \includegraphics[width=0.48\textwidth]{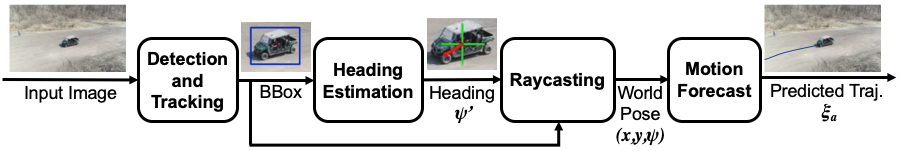}
    \caption{Vision sub-system. We detect and track the actor's bounding box, estimate its heading, and project its pose to world coordinates. A Kalman Filter predicts the actor's forecasted trajectory $\Path{a}$.}
    \label{fig:vision_pipeline}
\end{figure}

% The vision subsystem is core to enabling the system to localize and tracking moving targets with no additional setup needed. In this section, we present a vision pipeline to provide a predicted trajectory of the actor $\Path{a}$, defined in Section \ref{sec:problem_formulation}, which informs the trajectory cost function detailed in Section \ref{subsec:approach_planning}.
% accurate actor pose in the world frame. 
% Unlike previous works that operate either with high-accuracy indoor motion capture systems or use precision RTK GPS for actor localization,
We use only monocular images from the UAV's gimbal and the UAV's own state estimation to forecast the actor's trajectory in the world frame. The vision sub-system counts with four main steps:
% To enable real-time actor trajectory estimation, we propose first 
actor bounding box detection and tracking, heading angle estimation; global ray-casting, and finally a filtering stage. Figure \ref{fig:vision_pipeline} summarizes the pipeline.

% (3) Coupled with the bounding box, the image-space heading angle is then fed into the raycasting module to output the actor's current pose (i.e. position and heading) in the world frame. 
% Finally, (4) a Kalman Filter uses the current pose estimate to output a predicted actor trajectory,  shows the vision pipeline.
% \chnote{\textbf{Paragraph 1 : Overview - Motivation, what we have advanced, introduce detection, heading estimation, ray-casting}}

% In the context of aerial filming of a moving actor, accurate heading estimation is vital to delivering the director's artistic objectives and enabling aesthetic shots. Specifically, accurate heading estimation
% will enable better execution of the director's artistic objectives and more aesthetic 

% \chnote{we are moving towards a generalized platform, with minimal setup needed}
% \rbnote{explain why heading direction is important; get a bunch of content from ICRA paper}
% \chnote{introduce variables, and formal definition - semi-supervised heading estimation}
\begin{figure}[t]
    \center
    \includegraphics[width=0.44\textwidth]{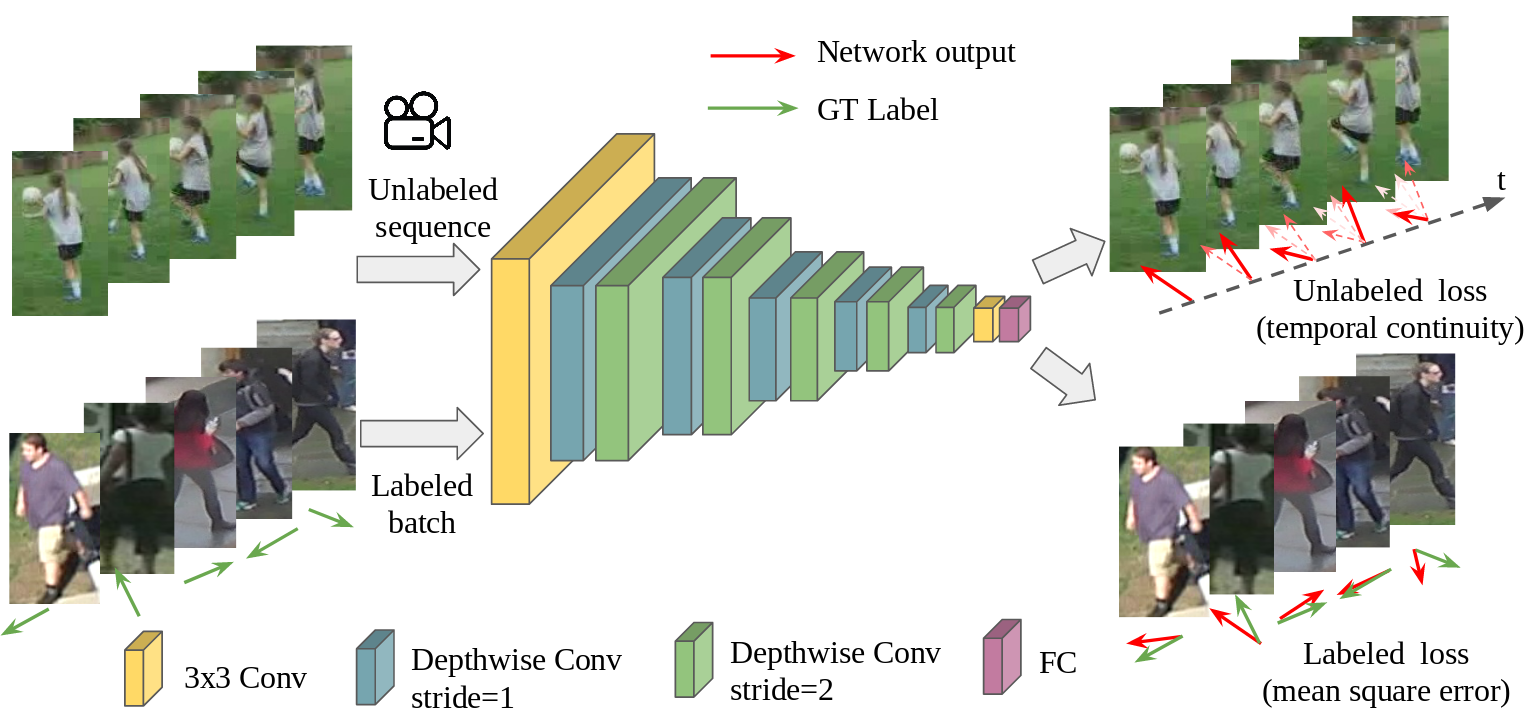}
    \caption{Heading detection network structure, from \cite{wang2019heading}. We leverage temporal continuity between frames to train a heading direction regressor with a small labeled dataset.}
    \label{fig:vision}
\end{figure}

\textit{a) Detection and tracking}:
% Using the monocular images from the camera, the detection module outputs a bounding box to initialize the tracking process, which operates at a higher frame rate. We then use a PD controller on the UAV's 3-axis camera gimbal to place the actor in the desired image position, as specified by the director's artistic guidelines (e.g., rule of thirds vs. centered).
Our detection module is based on the MobileNet network architecture, due to its low memory usage and fast inference speed, which are well-suited for real-time applications on an onboard computer. We use the same network structure as detailed in our previous work \cite{wang2019heading}. Our model is further trained with COCO~\cite{lin2014microsoft} and fine-tuned on a custom aerial filming dataset. We limit the detection categories to \textit{person}, \textit{car}, \textit{bicycle}, and \textit{motorcycle}, which commonly appear in aerial filming. After a successful detection we use  Kernelized Correlation Filters \cite{henriques2015high} to track the template over the next incoming frames. We actively position the independent camera gimbal with a PD controller to frame the actor on the desired screen position, following the commanded artistic principles (Fig.~\ref{fig:cine_params}).

\textit{b) Heading Estimation:}
Accurate heading angle estimation is vital for the UAV to follow the correct shot type (front, back, left, right). 
As discussed in \cite{wang2019heading}, human 2D pose estimation has been widely studied \cite{toshev2014deeppose, cao2017realtime}, but 3D heading direction cannot be trivially recovered directly from 2D points because depth remains undefined. 
Therefore, we use the model architecture from \cite{wang2019heading} (Fig. \ref{fig:vision}), which takes as input a bounding box image, and outputs the cosine and sine values of the heading angle. This network uses a double loss during training, summing both errors in heading direction and temporal continuity. The latter loss is particularly useful to train the regressor on small datasets, following a semi-supervised approach.

\textit{c) Ray-casting:}
Given the actor's bounding box and heading estimate on image space, we project the center-bottom of the bounding box onto the world's ground plane and transform the actor's heading using the camera's state estimation to obtain the actor's position and heading in world coordinates.

% Given the detected bounding box and image-space heading estimate from previous modules, we use raycasting to calculate the actor's pose in the world frame. The actor is assumed to be on a flat, horizontal ground plane. The raycasting module projects the bottom center of the detected bounding box onto the ground plane given the camera's known extrinsic and intrinsic parameters, and the global state estimate from the flight controller. The image-space heading estimation is similarly projected to the world frame.
% We use raycasting to calculate the actor's pose in the world frame, given the detected bounding box and the image-space heading estimate, from previously-described vision modules. 

\textit{d) Motion Forecasting:} The current actor pose updates a Kalman Filter (KF) to forecast the actor's trajectory $\Path{a}$. We use separate KF models for people and vehicle dynamics.

\subsection{Mapping sub-system}
\label{subsec:approach_mapping}

As explained in Section \ref{sec:problem_formulation}, the motion planner uses signed distance values $\map$ in the optimization cost functions. The role of the mapping is to register LiDAR points from the onboard sensor, update the occupancy grid $\grid$, and incrementally update the signed distance $\map$: 

\textit{a) LiDAR registration:} The laser at the bottom of the aircraft outputs roughly $300,000$ points per second. We register points in world coordinates using a rigid body transform between sensor and UAV, plus the UAV's state estimation, which fuses GPS, barometer, internal IMUs and accelerometers. 
% We also store each corresponding sensor coordinate, used next for the occupancy grid update. 
% Other types of sensors such as stereo pairs can replace this module if a similar output of hits and misses is provided.

\small
\begin{algorithm}[ht]
\caption{\small Update $\grid$ ($p_{sensor}$, $p_{point}$, $is\_hit$)}
\label{alg:ray}
\SetAlgoLined
% \KwResult{Write here the result }
 % $\Path{q} \gets \Path{q} {}_{previous}$\;
 Initialize $V^{change}_{occ} = \{\}$, $V^{change}_{free} = \{\}$ \\ 
 Initialize $l_{free}$, $l_{occ}$ \Comment{log-odds updates}\
 \For{\textrm{\textbf{each}} voxel $v$ between $p_{sensor}$ and $p_{point}$}{
  $v \gets v - l_{free}$\;
  \If{$v$ was occupied or unknown and now is free}{
   Append($v$, $V^{change}_{free}$)\;
   \For{\textrm{\textbf{each}} unknown neighbor $v_{unk}$ of $v$}{
    Append($v_{unk}$, $V^{change}_{occ}$)
   }
  }
  \If{$v$ is the endpoint and $is\_hit$ is true}{
   $v \gets v + l_{occ}$\;
   \If{$v$ was free or unknown and now is occupied}{
   Append($v$, $V^{change}_{occ}$)
   }
  }
 }
 \textbf{return} $V^{change}_{occ}$, $V^{change}_{free}$
\end{algorithm}
\normalsize

\textit{b) Occupancy grid $\grid$ update:} 
% \chnote{mention explicitly earlier that you're outputting obstacle updates, not building grid} 
% Points can represent either a hit (successful return) or a miss (return too far or too close from the sensor). We filter all expected misses caused by the aircraft's structure, and then probabilistically update all voxels from $\grid$ between the sensor and LiDAR point. In practice, we use a grid size of $250\times250\times100 m$, with $1 m$ square voxels that store an integer value between $0-255$ as the occupancy probability. Algorithm~\ref{alg:ray} covers the update process.
We use a grid size of $250\times250\times100 m$, with $1 m$ square voxels that store an $8$-bit integer value between $0-255$ (free - occupied) as the occupancy probability. 
All cells are initialized with $127$ (unknown).  
Algorithm~\ref{alg:ray} covers the grid update process.
The inputs to the algorithm are the sensor position $p_{sensor}$, the LiDAR point $p_{point}$, and a flag $is\_hit$ that indicates whether the point is a hit or miss. The endpoint voxel of a hit will be updated with log-odds value $l_{occ}$, and all cells in between sensor and endpoint will be updated by subtracting value $l_{free}$. We assume that all misses are returned as points at the maximum sensor range, and in this case only the cells between endpoint and sensor are updated.
Voxel state changes to \textit{occupied} or \textit{free} are stored in lists $V^{change}_{occ}$ and $V^{change}_{free}$. 
State changes are used for the signed distance update.
% \scnote{This algorithm seems standard occ grid mapping alg. Can we simply cite prob robotics and instead emphasize that you are tracking changed cells.}

% \small
% \begin{algorithm}[ht]
% \label{alg:ray}
% \SetAlgoLined
% % \KwResult{Write here the result }
%  % $\Path{q} \gets \Path{q} {}_{previous}$\;
%  Initialize $V^{change}_{occ}$, $V^{change}_{free}$ \Comment{changed voxels}\ 
%  Initialize $U_{free}$, $U_{occ}$ \Comment{probabilistic updates}\
%  \For{\textrm{\textbf{each}} voxel $V$ between $p_{sensor}$ and $p_{point}$}{
%   $V \gets V - U_{free}$\;
%   \If{$V$ was occupied or unknown and now is free}{
%    Append($V^{change}_{free}$,$V$)\;
%    \For{\textrm{\textbf{each}} unknown neighbor $V_{unk}$ of $V$}{
%     Append($V^{change}_{occ}$,$V_{unk}$)
%    }
%   }
%   \If{$V$ is the endpoint and $is\_hit$ is true}{
%    $V \gets V + U_{occ}$\;
%    \If{$V$ was free or unknown and now is occupied}{
%    Append($V^{change}_{occ}$,$V$)
%    }
%   }
%  }
%  \caption{\small Update $\grid$ ($p_{sensor}$, $p_{point}$, $is\_hit$)}
%  \textbf{return} $V^{change}_{occ}$, $V^{change}_{free}$
% \end{algorithm}
% \normalsize

\textit{c) Incremental distance transform $\map$ update:} 
We use the list of voxel state changes as input to an algorithm, modified from \cite{cover2013sparse}, that calculates an incremental truncated signed distance transform (iTSDT), stored in $\map$.
% We modify the algorithm presented in details by , which incrementally  
The original algorithm described by \cite{cover2013sparse} initializes all voxels in $\map$ as free, and as voxel changes arrive in sets $V^{change}_{occ}$ and $V^{change}_{free}$, it incrementally updates the distance of each free voxel to the closest occupied voxel using an efficient wavefront expansion technique within some limit (therefore truncated). Our problem, however, requires a \textit{signed} version of the DT, where the \textit{inside} and \textit{outside} of obstacles must be identified and given opposite signs. 
The concept of regions inside and outside of obstacles cannot be captured by the original algorithm, which provides only a iTDT (no sign).
Therefore, we introduced two important modifications:

\textit{i) Using obstacle borders.} 
% The original algorithm uses only the occupied cells of $\grid$, which are incrementally pushed into $\map$ using set $V^{change}_{occ}$. 
% We, instead, define the concept of \textit{obstacle border} cells, and push them incrementally as $V^{change}_{occ}$. 
% Let $v_{border}$ be an obstacle border voxel, and $V_{border}$ be the set of all border voxels in the environment. 
We define a border voxel $v_{border}$ as any voxel that is either a direct hit from the LiDAR (lines $13-15$ of Alg.~\ref{alg:ray}), or as any \textit{unknown} voxel that is a neighbor of a \textit{free} voxel (lines $5-9$ of Alg.~\ref{alg:ray}). In other words, the set $V_{border}$ will represent all cells that separate the known free space from unknown space in the map, whether this unknown space is part of cells inside an obstacle or cells that are actually free but just have not yet been cleared by the LiDAR. Differently from \cite{cover2013sparse}, our $\map$ uses updates $V^{change}_{occ}$ and $V^{change}_{free}$ to maintain the distance of any voxel to its closest border instead of the distance to the closest hit.
% By incrementally pushing  into , its data structure will maintain the current set of border cells $V_{border}$.
% By using the same algorithm described in  but now with this distinct type of data input, we can obtain the distance of in $\map$ to the closest obstacle border. One more step is required to obtain the sign of this distance.

\textit{ii) Querying $\grid$ for the sign.}
% The data structure of $\map$ only stores the distance of each cell to the nearest obstacle border. Therefore 
We query the value of $\grid$ to attribute the sign of the iTSDT, marking free voxels as positive, and unknown or occupied voxels as negative.

\subsection{Planning sub-system}
\label{subsec:approach_planning}

We want trajectories which are smooth, capture high quality viewpoints, avoid occlusion and are safe. Given the real-time nature of our problem, we desire fast convergence to locally optimal solutions rather than globally optimality taking a long time to obtain a solution. A popular approach is to cast the problem as an unconstrained optimization and apply covariant gradient descent~\cite{zucker2013chomp}. This is a quasi-Newton approach where some of the objectives have analytic Hessians that are easy to invert and are well-conditioned. Hence such methods exhibit fast convergence while being stable and computationally inexpensive. 
% \chnote{do we build on this then, not clear}

For this implementation, we use a waypoint parameterization of trajectories, i.e., $\Path{} \in \real^{n \times 3}$. The heading dimension $\psi(t)$ is set to always point the drone from $\Path{q}(t)$ towards $\Path{a}(t)$. We design a set of differentiable cost functions as follows: 

\paragraph{Smoothness}
We measure smoothness as the cumulative derivatives of the trajectory. Let $D$ be a discrete difference operator. The smoothness cost is:

\begin{equation}
	\begin{aligned}
		&\costFnSmooth{\Path{q}} = \frac{1}{t_{f}} \frac{1}{2} \int_0^{t_{f}} \sum_{d=1}^{d_{max}} \alpha_n (D^d \Path{q}(t))^2 dt \\
		&\approx \frac{1}{2(n-1)} Tr(\Path{q}^T A_{smooth} \Path{q} + 2\Path{q}^T b_{smooth} + c_{smooth}) \\
%&\nabla_{\Path{q}} \costFnSmooth{\Path{q}} = \frac{1}{n-1}(A_{smooth}\Path{q}+b_{smooth})
	\end{aligned}
\end{equation}
where $\alpha_n$ is a weight for different orders, and $d_{max}$ is the number of orders. We set $\alpha_n = 1, d_{max} = 3$. Note that  smoothness is a quadratic objective where the Hessian $A_{smooth}$ is analytic. 

\paragraph{Shot quality}

\begin{figure}[!t]
    \centering
    \includegraphics[width=0.5\textwidth]{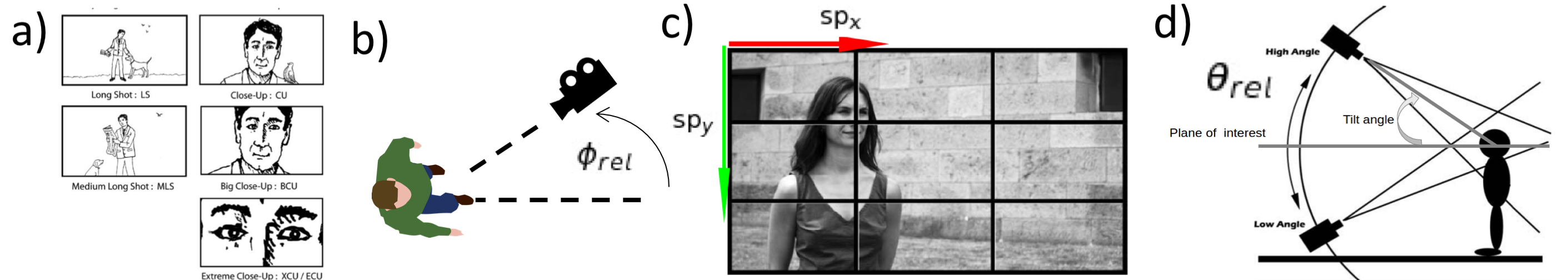}
    \caption{
    Shot parameters for shot quality cost function, adapted from \cite{bowen2013grammar}: a) shot scale $\rho$ corresponds to the size of the projection of the actor on the screen; b) line of action angle $\phi_{rel} \in [0,2\pi]$; c) screen position of the actor projection $sp_x,sp_y \in [0,1]$; d) tilt angle $\theta_{rel} \in [-\pi,\pi]$ 
    %\rbnote{modify the problem def picture to include besides artistic principles also environment generation and actor estimation. or is this figure redundant because all these things are shown in other figures like the cartoon of 1a and the map generation and actor localization in approach later on? maybe take out this fig}
    }
    \label{fig:cine_params}
\end{figure}

Written in a quadratic form, shot quality measures the average squared distance between $\Path{q}$ and an ideal trajectory $\Path{shot}$ that only considers positioning via cinematography parameters. $\Path{shot}$ can be computed analytically: for each point $\Path{a}(t)$ in the actor motion prediction, the ideal drone position lies on a sphere centered at $\Path{a}(t)$ with radius $\rho$ defined by the shot scale, relative yaw angle $\phi_{rel}$ and relative tilt angle $\theta_{rel}$ (Fig.~\ref{fig:cine_params}): 

\begin{equation}
	\begin{aligned}
		&\Path{shot}(t) = \Path{a}(t) + \rho 
		\begin{bmatrix}
		 cos(\psi_{a} + \phi_{rel}) sin(\theta_{rel})\\
		 sin(\psi_{a} + \phi_{rel}) cos(\theta_{rel})\\
		 cos(\theta_{rel})
		\end{bmatrix}\\
	\end{aligned}
\end{equation}

\begin{equation}
	\begin{aligned}
		&\costFnShot{\Path{q},\Path{a}} = \frac{1}{t_{f}} \frac{1}{2} \int_0^{t_{f}} ||\Path{q}(t)-\Path{shot}(\Path{a}(t))||^2 dt \\
		&								 \approx \frac{1}{2(n-1)} Tr(\Path{q}^T A_{shot} \Path{q} + 2\Path{q}^T b_{shot} + c_{shot}) \\
%&\nabla_{\Path{q}} \costFnShot{\Path{q},\Path{a}} = \frac{1}{n-1}(A_{shot}\Path{q}+b_{shot})
	\end{aligned}
\end{equation}
% \vspace{-0.0mm}
% \small{
% \begin{center} 
% $\costFnShot{\Path{q},\Path{shot}} = \frac{1}{t_{f}} \frac{1}{2} \int_0^{t_{f}} ||\Path{q}(t)-\Path{shot}(t)||^2 dt \approx \frac{1}{2(n-1)} Tr(\Path{q}^T A_{shot} \Path{q} + 2\Path{q}^T b_{shot} + c_{shot})$\\
% $\nabla \costFnShot{\Path{q}} = \frac{1}{n-1}(A_{shot}\Path{q}+b_{shot})$ 
% \end{center}
% }
% \vspace{-0.0mm}
% \normalsize

\paragraph{Safety}

Given the online map $\grid$, we calculate the TSDT $\map: \real^3 \to \real$ as described in Section~\ref{subsec:approach_mapping}. We adopt a cost function from \cite{zucker2013chomp} that penalizes proximity to obstacles:
\begin{equation}
c(p) = 
    \begin{cases}
      -\map(p)+\frac{1}{2} \epsilon_{obs} & \map(p)<0\\
      \frac{1}{2\epsilon_{obs}}(\map(p)-\epsilon_{obs})^2 & 0<\map(p)\leq\epsilon_{obs}\\
      0&\text{otherwise}
    \end{cases}
\end{equation}

We can then define the safety cost function~\cite{zucker2013chomp}:
\begin{equation}
    \costFnObs{\Path{q}, \mathcal{M}} = \int_{t=0}^{t_{f}} c(\xi(t)) \norm{\frac{d}{dt}\xi(t)}{} dt
\end{equation}

\paragraph{Occlusion avoidance}

Even though the concept of occlusion is binary, \textit{i.e}, we either have or don't have visibility of the actor, a major contribution of our past work \cite{bonatti2018autonomous} was defining a differentiable cost that expresses a viewpoint's occlusion intensity among arbitrary obstacle shapes.
% Formalizing the concept of occlusion cost for arbitrary obstacle shapes is , and has not been addressed in any prior work in the field for generic obstacle shapes.
Mathematically, we define occlusion as the integral of the TSDT cost $c$ over a 2D manifold connecting both trajectories $\Path{q}$ and $\Path{a}$. The manifold is built by connecting each drone-actor position pair in time using the path $p(\tau) = \tau\Path{q}(t) + (1-\tau)\Path{a}(\Path{a})$.

\begin{equation}
\begin{aligned}
&\costFnOcc{\Path{q},\Path{a},\mathcal{M}}\\
& = \int_{t=0}^{t_{f}} \left( \int_{\tau=0}^{1}\; c(p(\tau)) \norm{\frac{d}{d\tau} p(\tau)}{} d\tau \right)\, \norm{\frac{d}{dt}\Path{q}(t)}{} dt,
\end{aligned}
\end{equation}

Our objective is to minimize the total cost function $\costFn{\Path{q}}$ (Eq.~\ref{eq:main_cost}). We do so by covariant gradient descent, using the gradient of the cost function $\nabla J(\Path{q})$, and an analytic approximation of the Hessian $\nabla^2 J(\Path{q}) \approx (A_{smooth} + \lambda_1 A_{shot})$:
% This has to be inverted only once outside at the beginning of the optimization. The update rule is 
\begin{equation}
  \Path{q}^+ = \Path{q} - \frac{1}{\eta} (A_{smooth} + \lambda_1 A_{shot})^{-1} \nabla J(\Path{q})
\end{equation}
This step is repeated till convergence. We follow conventional stopping criteria for descent algorithms, and limit the maximum number of iterations. Note that we only perform the matrix inversion once, outside of the main optimization loop, rendering good convergence rates \cite{bonatti2018autonomous}. We use the current trajectory as initialization for the next planning problem.

\section{Experiments} 
\label{sec:experiments}

% In this section we detail our setup, and describe experiments that evaluate the performance of our system.

% !TEX root = ../root.tex

\subsection{Experimental setup}
\label{subsec:experiments_setup}
\begin{figure}[t]
    \centering
    \includegraphics[width=0.3\textwidth]{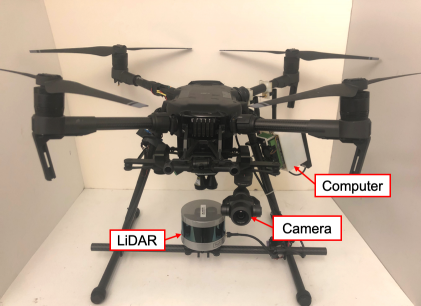}
    \caption{System hardware: DJI M210 drone, Nvidia TX2 computer, VLP16 LiDAR and Zenmuse X4S camera gimbal.}
    \label{fig:hardware}
\end{figure}

Our platform is a DJI M210 drone, shown in Figure \ref{fig:hardware}. All processing is done with a 
% To handle the computation of the vision, mapping and planning pipeline, we use as the onboard computer a 
NVIDIA Jetson TX2, with 8GB of RAM and 6 CPU cores. An independently controlled gimbal DJI Zenmuse X4S records high-resolution images. Our laser sensor is a Velodyne Puck VLP-16 Lite, with $\pm15^o$ vertical field of view and 100m max range. 
% \rbnote{update with new system setup. put a nice picture of the system with the velodyne}

\begin{figure}[!hb]
    \centering
    \includegraphics[width=0.3\textwidth]{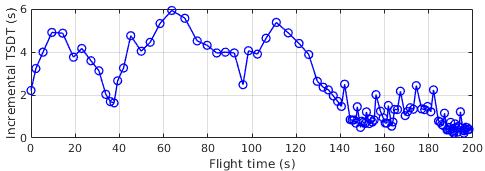}
    \caption{Incremental distance transform compute time over flight time. The first operations take significantly more time because of our map initialization scheme where all cells are initially considered as unknown instead of free. After the first minutes of flight incremental mapping is significantly faster.}
    \label{fig:dt_time}
\end{figure}

\subsection{Field test results}
\label{subsec:experiments_field_test}

\begin{figure}[b]
    \centering
    \includegraphics[width=0.45\textwidth]{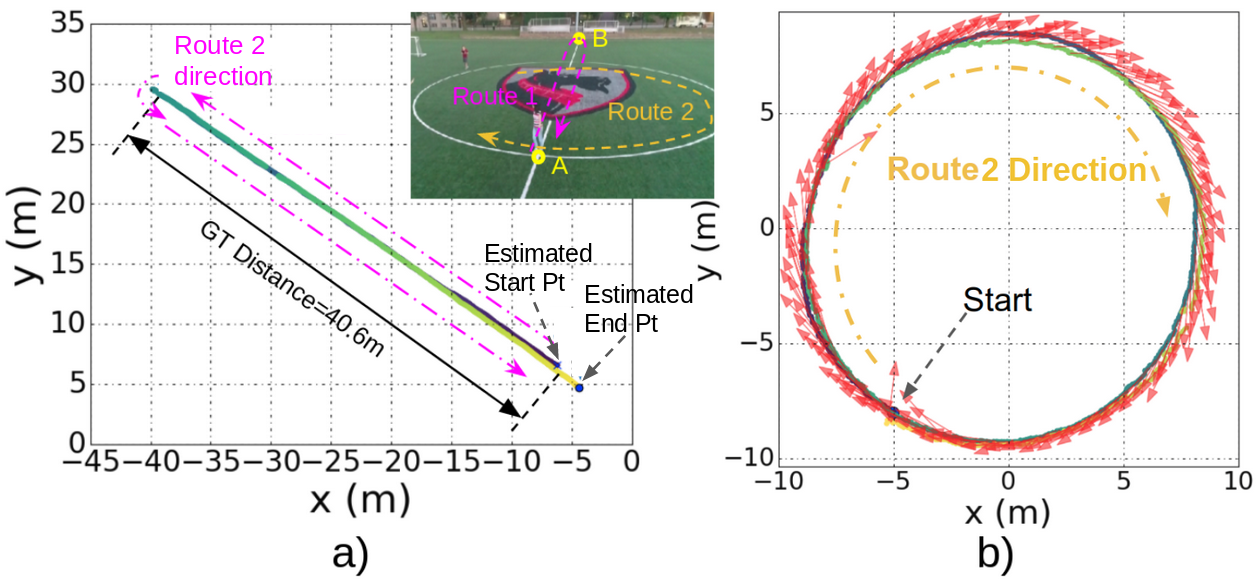}
    \caption{Pose and heading estimation results. a) Actor walks on a straight line from points A-B-A. Ground-truth trajectory length is 40.6m, while the estimated motion length is 42.3m. b) The actor walks along a circle. Ground-truth diameter is 18.3m, while the estimated diameter from ray casting is 18.7m. Heading estimation appears tangential to the ground circle.}
    \label{fig:vision_results}
\end{figure}

% \rbnote{vision results: show here the figure for the raycasting experiment on the football field that contains ground truth of raycasting versus what was actually performed}
\paragraph{Visual actor localization}
We validate the precision of our pose and heading estimation modules in two experiments where the drone hovers and visually tracks the actor. First, the actor walks between two points over a straight line, and we compare the estimated and ground truth path lengths. Second, the actor walks on a circle at the center of a football field, and we verify the errors in estimated positioning and heading direction. Fig.~\ref{fig:vision_results} summarizes our findings. 

% , we performed an experiment with the UAV system to evaluate vision performance against ground truth for the aerial filming application. I
% with a n the experiment, the actor walks around a marked circle twice, whilst keeping her heading tangent w.r.t. the circle, to showcase the accuracy of the estimated position and heading direction. Figure \ref{fig:vision_results} shows the results from the vision experiment. The results shows high consistency to ground truth...\chnote{are there quantitative results from this experiment?}

% \rbnote{vision results: show here the figure for the raycasting experiment on the football field that contains ground truth of raycasting versus what was actually performed}
% \begin{figure}[ht]
%     \centering
%     \includegraphics[width=0.5\textwidth]{figs/figs_icra_iros/ray_casting_exp.png}
%     \caption{Estimated actor trajectory by ray-casting module: a) The actor walks on a straight line from point A to B, and back to A again. While ground-truth trajectory length is 40.6m in total, the estimated motion length is of 42.3m; b) The actor walks twice on a circle. While the ground-truth diameter is 18.3m, the estimated diameter from ray casting is 18.7m. \chnote{maybe add clearer legend, what's GT distance}}
%     \label{fig:raycasting}
% \end{figure}

\begin{figure*}[ht]
    \center
    \includegraphics[width=1.0\textwidth]{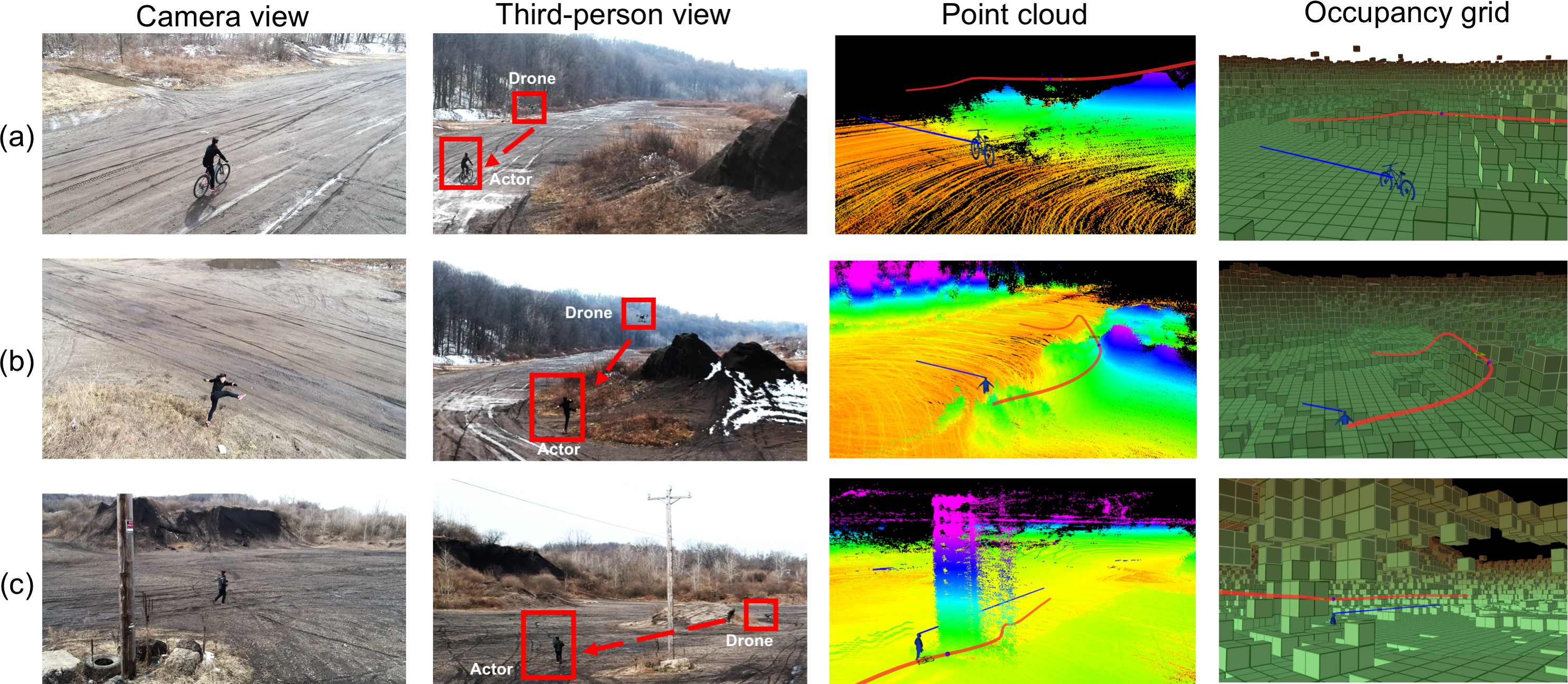}
    \caption{Field results: a) side shot following biker, b) circling shot around dancer, c) side shot following runner. The UAV trajectory (red) tracks the actor's forecasted motion (blue), and stays safe while avoiding occlusions from obstacles. We display accumulated point clouds of LiDAR hits and the occupancy grid. Note that LiDAR registration is noisy close to the pole in row (c) due to large electromagnetic interference of wires with the UAV's compass.}
    \label{fig:exp_main}
\end{figure*}

\paragraph{Integrated field experiments} We test the real-time performance of our integrated system in several field test experiments. We use our algorithms in unknown and unstructured environments outdoors, following different types of shots and tracking different types of actors (people and bicycles) at both low and high speeds in unscripted scenes. Fig.~\ref{fig:exp_main} summarizes the most representative shots, and the supplementary video 
\small \href{https://youtu.be/ookhHnqmlaU}{(https://youtu.be/ZE9MnCVmumc) }\normalsize
shows the final footages along with visualizations of point clouds and of the online map.

% \rbnote{planning results: talk about field experiments. show nice figure with results from person circular, right shots and bike right shot. don't forget the evidence grid, visual localization and heading estimation plots}

% \rbnote{probably won't be able to do the time-lapse figures like in ISER, but we can show other things that catch attention like the point clouds, evidence grid, and trajectories on RVIZ to give idea of time passing by}

% \begin{figure}[ht]
%     \centering
%     \includegraphics[width=0.4\textwidth]{figs/failure_case}
%     \caption{Failure case because of odom drift}
%     \label{fig:raycasting}
% \end{figure}

\paragraph{System statistics} We summarize runtime statistics in Table~\ref{tab:statistics}, and discuss online mapping details in Fig.~\ref{fig:dt_time}. While the vision networks takes up a large part of the system's RAM, CPU usage is fairly balanced accross systems.

\begin{table}[t]
\caption{System statistics}
\begin{tabular}{l|llllll}
\label{tab:statistics}
 \textbf{\pbox{0.2cm}{System}} & \textbf{\pbox{1.0cm}{Module}}& \textbf{\pbox{2.0cm}{CPU\\Thread (\%)}} & \textbf{\pbox{2.0cm}{RAM (MB)}}  & \textbf{\pbox{2.0cm}{Runtime (ms)}} \\
\hline
  & Detection & 57 & 2160  & 145 \\

Vision  & Tracking & 24 & 25  & 14.4 \\

 & Heading & 24 & 1768  & 13.9 \\

  & KF & 8 & 80  & 0.207 \\

\hline

 & Grid & 22 & 48  & 36.8 \\

Mapping & TSDF & 91 & 810  & 100-6000 \\

  & LiDAR & 24 & 9 & NA  \\

\hline

Planning  & Planner & 98 & 789  & 198 \\

  & DJI SDK & 89 & 40  & NA \\

\hline

\end{tabular}
\end{table}

% \rbnote{add runtime for each module as well? how long it takes (ms) for each part to complete a cycle}

% \missingfigure[figwidth=0.5\textwidth]{Time needed for Distance Transform Update over Flight Time. Time decreases as less cells will be updated.}
% !TEX root = ../root.tex

\subsection{Performance comparison with full information knowledge}
\label{subsec:experiments_comparison}

\begin{figure}[t]
    \centering
    \includegraphics[width=0.45\textwidth]{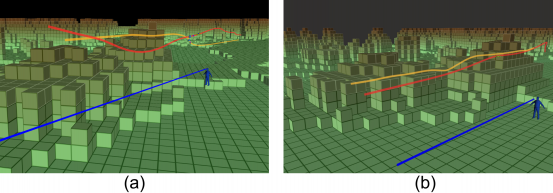}
    \caption{Performance comparisons. a) Planning with full knowledge of the map (yellow) versus with online mapping (red), displayed over ground truth map grid. Online map trajectory is less smooth due to a imperfect LiDAR registration and new obstacle discoveries as flight progresses. b) Planning with perfect ground truth of actor's location versus noisy actor estimate with artificial noise of 1m amplitude. The planner is able to handle noisy actor localization well due to smoothness cost terms, with final trajectory similar to ground-truth case.}
    \label{fig:comparison}
\end{figure}

An important hypothesis behind our system is that we can operate with insignificant loss in performance using noisy actor localization and a partially known maps. We compare our system with three assumptions from previous works:

\paragraph{Ground-truth obstacles vs. online map} We compare average planning costs between results from a real-life test where the planner operated while mapping the environment in real time with planning results with the same actor trajectory but with full knowledge of the map beforehand. Results are averaged over 140 s of flight and approximately 700 planning problems. Table~\ref{tab:comp} shows a small increase in average planning costs, and Fig~\ref{fig:comparison}a shows that qualitatively both trajectories differ minimally. The planning time, however, doubles in the online mapping case due to extra load on CPU.

\begin{table}[t]
\caption{Performance comparisons}
\begin{tabular}{l|llllll}
\label{tab:comp}
 \textbf{\pbox{2.5cm}{Planning Condition}} & \textbf{\pbox{2.5cm}{Avg. plan\\time(ms)}}& \textbf{\pbox{1.5cm}{Avg. cost}} & \textbf{\pbox{1.5cm}{Median cost}}  \\
\hline
Ground-truth map  & 32.1 & 0.1022 & 0.0603 \\
Online map  & 69.0 & 0.1102 & 0.0825 \\
\hline
Ground-truth actor  & 36.5 & 0.0539 & 0.0475 \\
Noise in actor  & 30.2 & 0.1276 & 0.0953 \\
\hline
\end{tabular}
\end{table}

\paragraph{Ground-truth actor versus noisy estimate} We compare the performance between simulated flights where the planner has full knowledge of the actor's position versus artificially noisy estimates with 1m amplitude. Results are also averaged over 140 s of flight and approximately 700 planning problems, and are displayed on Table~\ref{tab:comp}. The large cost difference is due to the shot quality cost, which relies on the actor's position forecast and is severely penalized by the noise. However, if compared with the actor's ground-truth trajectory, the difference in cost would be significantly smaller, as seen by the proximity of both final trajectories in Fig~\ref{fig:comparison}b. These results offer insight on the importance of our smoothness cost function when handling the noisy visual actor localization.

\paragraph{Height map assumption vs. 3D map} As seen in Fig.~\ref{fig:exp_main}c, our current system is capable of avoiding unstructured obstacles in 3D environments such as wires and poles. This capability is a significant improvement over our previous work \cite{bonatti2018autonomous}, which used a height map assumption.

\section{Conclusion}
\label{sec:conclusion}
% We present a complete autonomous cinematography platform that can visually localize actors and operate in a previously unmapped environment \chnote{aestheticness?}. By estimating actor pose directly from camera image and building an online map, we address two key limitations of previous approaches: the need for additional sensors to localize actors, and a prior map. We verify that our system produces results comparable to or above previous state-of-the-art approaches whilst removing the need to full map and actor knowledge. Finally, simulations and field experiments show the system’s robustness and real-time performance while tracking dynamic targets among unstructured 3D obstacles. \chnote{variety of actor types? different shot type} 
% We address two key limitations of current autonomous cinematographers: the need for additional sensors to localize actors, and a prior map. By estimating actor pose directly from camera image and builds an online map, which are then used to efficiently compute a trajectory that balances safety, motion smoothness, occlusion, and artistic guidelines. 
% We present a complete autonomous cinematography platform that can visually localize dynamic actors and operate in a previously unmapped environment to capture aesthetic shots. 

We present a complete system for autonomous aerial cinematography that can localize and track actors in unknown and unstructured environments with onboard computing in real time. Our platform uses a monocular visual input to localize the actor's position, and a custom-trained network to estimate the actor's heading direction. Additionally, it maps the world using a LiDAR and incrementally updates a signed distance map. Both of these are used by a camera trajectory planner that produces smooth and artistic trajectories while avoiding obstacles and occlusions. We evaluate the system extensively in different real-world tasks with multiple shot types and varying terrains.  

We are actively working on a number of directions based on lessons learned from field trials. Our current approach assumes a static environment. Even though the our mapping can tolerate motion, a principled approach would track moving objects and forecast their motion. The TSDT is expensive to maintain because whenever unknown space is cleared, a large update is computed. We are looking into a just-in-time update that processes only the subset of the map queried by the planner, which is often quite small. 

Currently, we do not close the loop between the image captured and the model used by the planner. Identifying model errors, such as actor forecasting or camera calibration, in an online fashion is a challenging next step. The system may also lose the actor due to tracking failures or sudden course changes. An exploration behavior to reacquire the actor is essential for robustness.

\section*{ACKNOWLEDGMENT}

We thank Mirko Gschwindt, Xiangwei Wang, Greg Armstrong for the assistance in field experiments and robot construction.

%%%%%%%%%%%%%%%%%%%%%%%%%%%%%%%%%%%%%%%%%%%%%%%%%%%%%%%%%%%%%%%%%%%%%%%%%%%%%%%%

% \bibliographystyle{IEEEtran}
% \bibliography{IEEEabrv,IEEEexample}
\footnotesize{
\bibliographystyle{IEEEtran}
\bibliography{IEEEabrv,IEEEexample}
}

\end{document}